\documentclass[runningheads]{llncs}

\usepackage{eccv}

\usepackage{eccvabbrv}

\usepackage{graphicx}
\usepackage{booktabs}

\usepackage[accsupp]{axessibility}  
\usepackage{multirow}  
\usepackage{tabularx} 

\usepackage[pagebackref,breaklinks,colorlinks,citecolor=eccvblue]{hyperref}

\usepackage{orcidlink}

\begin{document}

\title{TableVision: A Large-Scale Benchmark for Spatially Grounded Reasoning over Complex Hierarchical Tables} 

\titlerunning{TableVision: A Large-Scale Benchmark for Table Reasoning}

\author{
    Xiaoyu Chen\inst{1}\thanks{Equal contribution} \and
    Lu Dai\inst{1,2}\protect\footnotemark[1] \and
    Hanqing Wang\inst{1} \and
    Zhuoyu Li\inst{3} \and
    Wenbin Dai\inst{4} \and
    Yanzong Zheng\inst{1} \and
    Zhenggang Xia\inst{1} \and
    Junyong Lin\inst{1} \and
    Hui Xiong\inst{1,2}\thanks{Corresponding author}
}

\authorrunning{X.~Chen, L.~Dai, et al.}

\institute{
    The Hong Kong University of Science and Technology (Guangzhou) \and
    The Hong Kong University of Science and Technology \and
    The Chinese University of Hong Kong \and
    Shanghai Jiao Tong University \\
    \email{xchen786@connect.hkust-gz.edu.cn, xionghui@ust.hk} \\
    \vspace{0.5em}
}

\maketitle
\begin{center}
    \captionsetup{type=figure}
    \includegraphics[width=\linewidth]{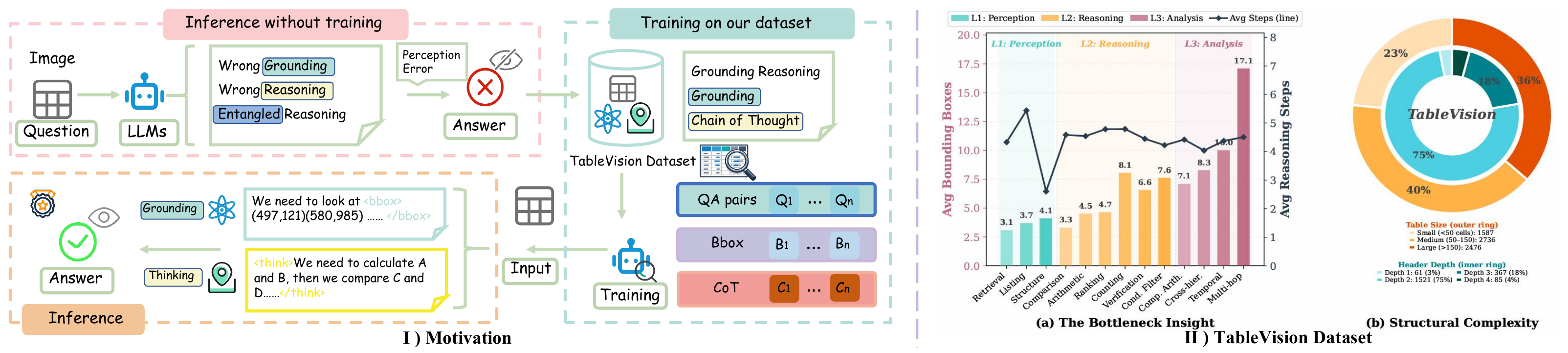}
    \caption{\textbf{The Motivation and Insight of TableVision.} 
\textbf{(I) Motivation:} Traditional MLLMs often suffer from perception errors due to entangled grounding and reasoning (top left). Our decoupled approach (bottom left) leverages the TableVision dataset to physically anchor the reasoning chain via explicit spatial constraints. 
\textbf{(II) Dataset Insight:} (a) The Bottleneck Insight reveals that as task complexity increases (L1 $\to$ L3). (b) Structural Complexity highlights the diverse distribution of table sizes and header depths in our dataset.}
\end{center}
\label{fig:teaser}

\begin{abstract}
Structured tables are essential for conveying high-density information in professional domains such as finance, healthcare, and scientific research. Despite the progress in Multimodal Large Language Models (MLLMs), reasoning performance remains limited for infographics, especially images of complex tables with hierarchical layouts. In this paper, we identify a critical \textbf{Perception Bottleneck} through quantitative analysis. We find that as task complexity scales, the number of involved discrete visual regions increases disproportionately. This processing density leads to an internal ``Perceptual Overload", where MLLMs struggle to maintain accurate spatial attention during implicit generation. To address this bottleneck, we introduce \textbf{TableVision}, a large-scale, trajectory-aware benchmark designed for spatially grounded reasoning. TableVision stratifies tabular tasks into three cognitive levels (\textit{Perception, Reasoning, and Analysis}) across 13 sub-categories. By utilizing a rendering-based deterministic grounding pipeline, the dataset explicitly couples multi-step logical deductions with pixel-perfect spatial ground truths, comprising 6,799 high-fidelity reasoning trajectories. Our empirical results, supported by diagnostic probing, demonstrate that explicit spatial constraints significantly recover the reasoning potential of MLLMs. Furthermore, our two-stage decoupled framework achieves a robust \textbf{12.3\% overall accuracy improvement} on the test set. TableVision provides a rigorous testbed and a fresh perspective on the interplay between perception and logic in document understanding.
\end{abstract}


\section{Introduction}
\label{sec:intro}

In domains such as finance, healthcare, and scientific research, high-density information is predominantly presented in structured tables. Traditional data processing paradigms typically rely on external OCR tools to serialize these tables into one-dimensional text (e.g., Markdown)~\cite{Kim2021OCRFreeDU}. However, this modality transformation is not only limited by the recognition ability of external tools, but also destroys the inherent 2D physical topological structure and spatial alignment of complex tables, especially those with hierarchical headers~\cite{singh2025lost, Xu2025SpatialBenchBM}. Recently, with the rapid development of Multimodal Large Language Models (MLLMs)~\cite{Achiam2023GPT4TR, Bai2025Qwen25VLTR, Dai2023InstructBLIPTG}, the research community has begun to explore unified and end-to-end methods for image-based table reasoning, which aims to preserve and leverage these 2D layouts within the native image space\cite{Cao2026OrthogonalHD}.

Despite the advantage of preserving structural information, current end-to-end MLLMs still struggle to deal with dense, complex tables~\cite{DeepSeekAI2025DeepSeekR1IR}. Complex tabular reasoning requires the model to simultaneously process dense intra-table visual information and execute multi-step logical computation during the implicit generation process. Through a quantitative characterization of task complexity, we identify a distinct decoupling in model requirements: while the reasoning depth remains consistent across levels, the perceptual demand scales significantly as task complexity increases. This pattern suggests that the primary bottleneck in complex table understanding is not necessarily logical depth, but the perceptual overload induced by multi-anchor grounding. This intensified density of visual information processing leads to an internal "Perceptual Overload": the model struggles to maintain accurate spatial attention when implicitly aligning massive visual elements, easily losing focus within the dense and topological grid\cite{singh2025lost}. Consequently, the absence of explicit 2D spatial grounding guidance constitutes the Perception Bottleneck that restricts the complex reasoning capabilities of current MLLMs\cite{kang2025vgentvisualgroundingmodular, Cheng2024SpatialRGPTGS, Janner2017GroundedSpatial}.
\label{sec:intro}
\begin{figure}[t]
    \centering
    \includegraphics[width=1\linewidth]{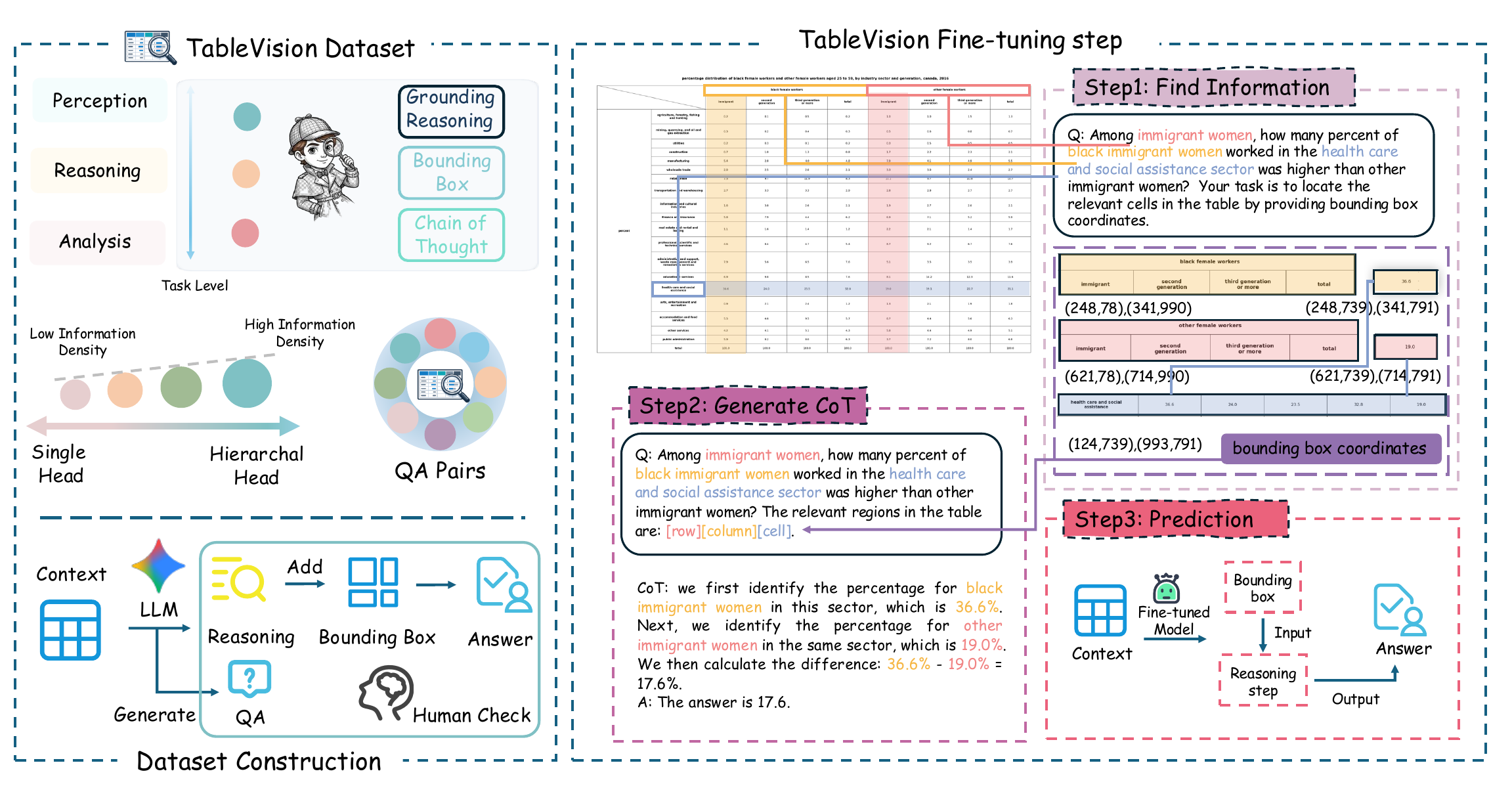}
    \caption{\textbf{Overview of the TableVision benchmark and the proposed grounding-enhanced reasoning framework.} Our TableVision dataset encompasses diverse multimodal reasoning tasks across complex tables with varying structural densities (e.g., hierarchical headers). To mitigate \textit{structural entanglement}, we introduce a decoupled fine-tuning paradigm. Rather than predicting the final answer end-to-end, the model explicitly performs question-aware structural localization (Step 1) to identify relevant cell coordinates. These spatial anchors are then integrated to condition the multi-step logical calculation (Step 2 \& 3), enabling highly reliable and grounded table reasoning.}
    \label{fig:structure}
\end{figure}

Existing tabular datasets mostly provide only final answers, lacking trajectory annotations that explicitly bind reasoning steps to underlying information pieces~\cite{Zhang2025T2RbenchAB, Chen2020OpenQA, Chen2019TabFactAL, Cheng2021HiTabAH, Gupta2020INFOTABSIO, Herzig2020TaPasWS}. To investigate and mitigate the perception bottleneck, it is essential to enhance spatial understanding in logical reasoning and provide models with fine-grained spatial alignment supervision. To this end, we construct the \textbf{TableVision} dataset, focusing on spatially grounded reasoning. To systematically evaluate the structured-table understanding capabilities of MLLMs, we stratify tabular question-answering tasks into three cognitive levels of varying difficulty (L1: Perception, L2: Reasoning, L3: Analysis), which are further subdivided into 13 sub-tasks. Furthermore, to minimize the alignment noise introduced by automated LLM or OCR annotations, we design a Rendering-Based Deterministic Grounding Pipeline. This pipeline anchors multi-step logical deductions directly to pixel-perfect spatial ground truths (bounding boxes), ultimately generating 6,799 high-fidelity multimodal reasoning trajectories.

To evaluate the performance of current multimodal models comprehensively, we conducted an extensive evaluation of 10 representative MLLMs on the proposed TableVision benchmark. The results reveal that existing models generally underperform on complex analysis tasks involving high visual loads. Subsequently, a diagnostic probing experiment demonstrates the potential for performance recovery: when provided with exact spatial coordinates (Oracle Grounding) as input prompts, the reasoning accuracy of baseline models on complex tables improves significantly. This empirical finding suggests that explicit spatial guidance can effectively activate the logical reasoning potential of MLLMs. Motivated by this observation, we implement a two-stage decoupled framework (Stage-1 Visual Localization $\rightarrow$ Stage-2 Reasoning Generation). Under a fully automated predictive setting, this pipeline achieves an overall accuracy improvement of 12.3\% across the test set, establishing a robust empirical baseline for multimodal structured tabular reasoning.

Our main contributions are summarized as follows:
\begin{itemize}
    \item \textbf{Identification of the Perception Bottleneck:} We quantitatively demonstrate that the increased density of intra-table visual retrieval, rather than the depth of logical deduction, is the core factor limiting the scalability of current end-to-end MLLMs on complex tabular tasks.
    \item \textbf{The TableVision Benchmark:} We release a tabular dataset comprising 6,799 high-fidelity reasoning trajectories. Spanning three cognitive levels, this benchmark explicitly couples multi-step logical reasoning with 2D physical bounding boxes to evaluate the spatial alignment capabilities of multimodal models.
    \item \textbf{Deterministic Construction Pipeline:} We provide a rendering-based alignment methodology that deterministically maps implicit logical derivations to image pixels, offering a reliable approach for constructing low-noise document datasets.
    \item \textbf{Empirical Analysis and Baseline:} Through extensive benchmarking and a two-stage decoupled experiment, we empirically validate the practical effectiveness of explicit spatial grounding supervision in enhancing the accuracy of multimodal structured reasoning.
\end{itemize}

\section{Related work}

\subsection{Multimodal Table Understanding}
Multi-modal Large Language Models (MLLMs), such as LLaVA \cite{liu2023visualinstructiontuning, xu2024llavauhdlmmperceivingaspect, liu2025llavaplus} and Qwen-VL \cite{wang2024qwen2vlenhancingvisionlanguagemodels, Bai2025Qwen25VLTR, bai2025qwen3vltechnicalreport} series, have advanced rapidly. To adapt them for table understanding, several end-to-end approaches implicitly learn layouts via data-driven alignment \cite{zheng2024multimodaltableunderstanding, ComTQA}, structural reinforcement learning \cite{kang2025grpoboostcomplexmultimodal}, or autoregressive bounding box prediction \cite{liu2025tellenhancingkeyinformation}. 

To circumvent the structural entanglement in end-to-end VLMs, other works adopt a modality conversion pipeline. These methods serialize table images into one-dimensional textual formats \cite{zhong2020imagebasedtablerecognitiondata, liu2023deplotoneshotvisuallanguage, yutong2025talenttablevqaaugmented}, utilize graph networks to model local relationships \cite{zhao2023localizeretrievefusegeneralized}, or transform tables into executable representations (e.g., Pandas or SQL) for program-aided computation \cite{cao2026tablemasterrecipeadvancetable, Cao_2023}. While effective, text-based serialization often sacrifices vital spatial topology, and program-aided pipelines risk error cascading on dense tabular structures.

\subsection{Think with Images}
Transcending the view of images as merely static inputs, recent multimodal reasoning advances towards a "Thinking with Images" paradigm \cite{su2025thinking}, actively utilizing visual representations as manipulable intermediate steps. Early agentic models dynamically invoke external tools (e.g., image cropping) to acquire fine-grained details \cite{zheng2025deepeyesincentivizingthinkingimages, hong2025deepeyesv2agenticmultimodalmodel}. However, physical cropping severs the global table topology and incurs high latency. Alternatively, data-centric and modular approaches explicitly align reasoning with synthetic bounding-box data or low-level grounding \cite{yang2025scalingtextrichimageunderstanding, kang2025vgentvisualgroundingmodular}. Further extensions treat models as visual programmers orchestrating external APIs \cite{gupta2022visualprogrammingcompositionalvisual, hu2023promptcappromptguidedtaskawareimage, mallis2025cadassistanttoolaugmentedvllmsgeneric} or enable closed-loop internal visual imagination \cite{chameleonteam2025chameleonmixedmodalearlyfusionfoundation, sun2024generativemultimodalmodelsincontext, xu2026visualplanningletsthink}. Despite these advancements, applying explicit spatial reasoning directly to structurally dense and semantically complex tables remains an open challenge.

\subsection{Table Reasoning Datasets}
The evolution of table reasoning datasets reflects a shift from text processing to multimodal visual understanding. Early benchmarks \cite{wtq, ToTTo, HybridQA, OTTQA, HiTab, Wu2024TableBenchAC, wu2025mmqa, OpenViTabQA} serialized tables into one-dimensional formats (e.g., HTML, Markdown). Although advancing semantic parsing, this dimensionality reduction strips away crucial spatial layouts, blinding models to original topological structures. 

To bridge this gap, hybrid datasets \cite{PubTabNet, TATQA, TabMWP, TableVQA, MMTU, RealHiTBench} incorporated both image and text inputs to tackle more complex, real-world scenarios. However, these datasets frequently treat images merely as auxiliary context or rely heavily on external OCR, compromising essential spatial alignment.

Consequently, recent focus has shifted towards purely vision-centric datasets \cite{SynthTabNet, WikiDT, ComTQA, fu2025refocus, VisualTableQA} for evaluating end-to-end MLLM perception. Yet, existing datasets predominantly offer only final answers or lack fine-grained coordinate annotations for intermediate reasoning. In contrast, our TableVision provides deterministic spatial supervision, explicitly coupling each logical reasoning step with bounding boxes to systematically activate MLLMs' multi-step visual reasoning potential.

\section{The TableVision Dataset}
\label{sec:dataset}
\subsection{Design Philosophy: Disentangling Perception from Reasoning}
\label{subsec:philosophy}

As compared in Table~\ref{tab:dataset_comparison}, existing tabular VQA datasets predominantly follow an implicit end-to-end paradigm, which often leads to spatial disorientation and the capture of incorrect semantic content during long-sequence generation. To mitigate this perception bottleneck, we introduce \textbf{TableVision}. Departing from direct prediction, we advocate for a decoupled \textbf{"Locate-then-Reason"} paradigm. Formally, each instance is defined as a quadruplet $\mathcal{T} = \{I, Q, \mathcal{S}, A\}$, where the core innovation lies in the \textbf{Spatial Evidence Set} $\mathcal{S}$. In this framework, the model is expected to first identify and extract key cell or header regions (bounding boxes $\mathcal{B}$) relevant to query $Q$ from image $I$. These precisely localized visual evidences are then re-incorporated as inputs, enabling the model to execute a logical Chain-of-Thought $C$ and derive the final answer $A$ while effectively bypassing visual noise. This design physically \textbf{disentangles} information localization from symbolic computation, allowing \textit{TableVision} to serve as a granular diagnostic tool to distinguish whether a failure stems from spatial mis-localization or logical error.

\begin{table}[t]
    \centering
    \caption{\textbf{Comprehensive comparison with existing tabular datasets.} "Hier." indicates multi-level hierarchical headers; "Bbox" denotes spatial grounding; "CoT" refers to step-by-step logical rationale. Compared to existing benchmarks, \textbf{TableVision} provides reasoning steps and bounding boxes that unify structural perception with complex multi-step reasoning.}
    \label{tab:dataset_comparison}
    \setlength{\tabcolsep}{1.5pt} 
    \scriptsize
    \begin{tabular}{@{} l c c c c l c r @{}}
        \toprule
        \textbf{Dataset} & \textbf{Year} & \textbf{Format} & \textbf{Header} & \textbf{Bbox} & \textbf{Task Type} & \textbf{CoT} & \textbf{Size} \\
        \midrule
        WTQ\cite{wtq} & 2015 & Text & Flat & $\times$ & Retr., Multi-hop & $\times$ & 220k QA \\
        ToTTo\cite{ToTTo} & 2020 & Text & Flat & $\times$ & Table-to-Text & $\times$ & 120k Tbl \\
        PubTabNet\cite{PubTabNet} & 2020 & Img & Flat & $\times$ & Structure Recog. & $\times$ & 568k Tbl \\
        HybridQA\cite{HybridQA} & 2020 & Text & Flat & $\times$ & Retr., Multi-hop & $\times$ & 70k QA \\
        OTT-QA\cite{OTTQA} & 2021 & Text & Flat & $\times$ & Retr., Multi-hop & $\times$ & 45k QA \\
        TAT-QA\cite{TATQA} & 2021 & Text & Flat & $\times$ & Retr., Arith. & $\times$ & 16k QA \\
        HiTab\cite{HiTab} & 2022 & Text & \textbf{Hier.} & $\times$ & Retr., Multi-hop & $\times$ & 10.6k QA \\
        SynthTabNet\cite{SynthTabNet} & 2022 & Img & Flat & $\checkmark$ & Structure Recog. & $\times$ & 600k Tbl \\
        TabMWP\cite{TabMWP} & 2023 & Text & Flat & $\times$ & Arith., Multi-hop & $\checkmark$ & 38k QA \\
        TableVQA-B.\cite{TableVQA} & 2024 & Img & Flat & $\times$ & Retr., Multi-hop & $\checkmark$ & 1.5k QA \\
        WikiDT\cite{WikiDT} & 2024 & Img & Flat & $\times$ & Extractive QA & $\times$ & 70k QA \\
        ComTQA\cite{ComTQA} & 2024 & Img & Flat & $\times$ & Retr., Multi-hop & $\times$ & 9k QA \\
        MMTU\cite{MMTU} & 2025 & Img & Flat & $\times$ & Retr., Multi-hop & $\times$ & 8.9k QA \\
        ReFocus\cite{fu2025refocus} & 2025 & Img & Flat & $\checkmark$ & Retr., Multi-hop & $\times$ & 14k QA \\
        TableBench\cite{Wu2024TableBenchAC} & 2025 & Text & Flat & $\times$ & Retr., Multi-hop & $\times$ & 886 QA \\
        MMQA\cite{wu2025mmqa} & 2025 & Text & Flat & $\times$ & Retr., Multi-hop & $\times$ & 3.3k QA \\
        RealHiTBench\cite{RealHiTBench} & 2025 & Img/Text & \textbf{Hier.} & $\times$ & Retr., Multi-hop & $\checkmark$ & 3.5k QA \\
        Visual-TQA\cite{VisualTableQA} & 2025 & Img & Flat & $\times$ & Retr., Multi-hop & $\times$ & 6k QA \\
        Open-ViTabQA\cite{OpenViTabQA} & 2025 & Text & Flat & $\times$ & Extractive QA & $\times$ & 9.9k QA \\
        \midrule
        \textbf{TableVision (Ours)} & 2026 & \textbf{Img} & \textbf{Hier.} & \boldmath$\checkmark$ & \textbf{Retr./Arith./Multi-hop} & \boldmath$\checkmark$ & \textbf{6.8k QA} \\
        \bottomrule
    \end{tabular}
\end{table}

\subsection{Data Taxonomy and Hierarchical Distribution}
\label{subsec:taxonomy}

To systematically benchmark multimodal tabular intelligence, we categorize tasks into three cognitive levels (L1 to L3) spanning 13 fine-grained operation types. This hierarchical structure enables a comprehensive evaluation of model performance across varying degrees of complexity. Our dataset comprises 6,799 decoupled reasoning instances, with 5,498 for training and 1,301 for testing. As summarized in Table~\ref{tab:dataset_distribution}, the visual perception burden (measured by the average number of bounding boxes) scales significantly across these levels.

\noindent\textbf{Level 1: Perception.}
L1 tasks focus on basic visual navigation and header mapping. It includes operations such as \textit{Retrieval}, \textit{Listing}, and \textit{Structure Identification}. These tasks require the model to locate specific cells or recognize the hierarchical layoutwithout complex numerical computation.

\noindent\textbf{Level 2: Reasoning.}
L2 tasks introduce logical operations based on successful localization. It encompasses six operation types: \textit{Comparison}, \textit{Arithmetic}, \textit{Ranking}, \textit{Conditional Filtering}, \textit{Counting}, and \textit{Verification}. These queries require the model to perform atomic reasoning after anchoring the relevant visual evidence. 

\noindent\textbf{Level 3: Analysis.}
L3 represents the highest complexity, involving multi-step structural traversals. Operations include \textit{Compositional Arithmetic}, \textit{Cross-hierarchy Aggregation}, \textit{Multi-hop Reasoning}, and \textit{Temporal Analysis}. L3 queries require models to concurrently manage and process a large density of visual anchors. Significant performance drops at this level highlight limitations in maintaining coherent spatial-logical associations over complex structural traversals.

\begin{table}[t]
\centering
\caption{\textbf{Hierarchical distribution and complexity metrics of TableVision.} The taxonomy stratifies queries into three cognitive levels. Notably, the visual perception burden (Avg Bbox) scales exponentially from simple Retrieval to complex Multi-hop analysis.}
\label{tab:dataset_distribution}
\setlength{\tabcolsep}{5pt} 
\scriptsize 
\begin{tabular}{@{} l l c c c c c @{}}
\toprule
\textbf{Cognitive Level} & \textbf{Category} & \textbf{Train} & \textbf{Test} & \textbf{Total} & \textbf{Avg Bbox} & \textbf{Avg Steps} \\ 
\midrule
\multirow{3}{*}{\textbf{L1: Perception}} 
& Retrieval & 1,813 & 371 & 2,184 & 3.10 & 4.33 \\
& Listing & 109 & 28 & 137 & 3.72 & 5.44 \\
& Structure & 394 & 114 & 508 & 4.12 & 2.61 \\
\midrule
\multirow{6}{*}{\textbf{L2: Reasoning}} 
& Comparison & 494 & 113 & 607 & 3.33 & 4.58 \\
& Arithmetic & 393 & 48 & 441 & 4.52 & 4.54 \\
& Ranking & 344 & 47 & 391 & 4.66 & 4.78 \\
& Counting  & 310 & 92 & 402 & 8.07 & 4.79 \\
& Cond. Filtering & 50 & 13 & 63 & 7.63 & 4.22 \\
& Verification & 363 & 112 & 475 & 6.57 & 4.45 \\
\midrule
\multirow{4}{*}{\textbf{L3: Analysis}} 
& Comp. Arithmetic & 363 & 107 & 470 & 7.12 & 4.42 \\
& Multi-hop & 350 & 104 & 454 & 17.11 & 4.51 \\
& Temporal & 296 & 80 & 376 & 10.02 & 4.37 \\
& Cross-hier. Agg. & 219 & 72 & 291 & 8.29 & 4.04 \\
\midrule
\textbf{Overall} & \textbf{13 Categories} & \textbf{5,498} & \textbf{1,301} & \textbf{6,799} & \textbf{5.79} & \textbf{4.33} \\
\bottomrule
\end{tabular}
\end{table}

\subsection{The Trajectory-Aware Construction Pipeline}
\label{subsec:pipeline}
\begin{figure}[t]
    \centering
    \includegraphics[width=1\linewidth]{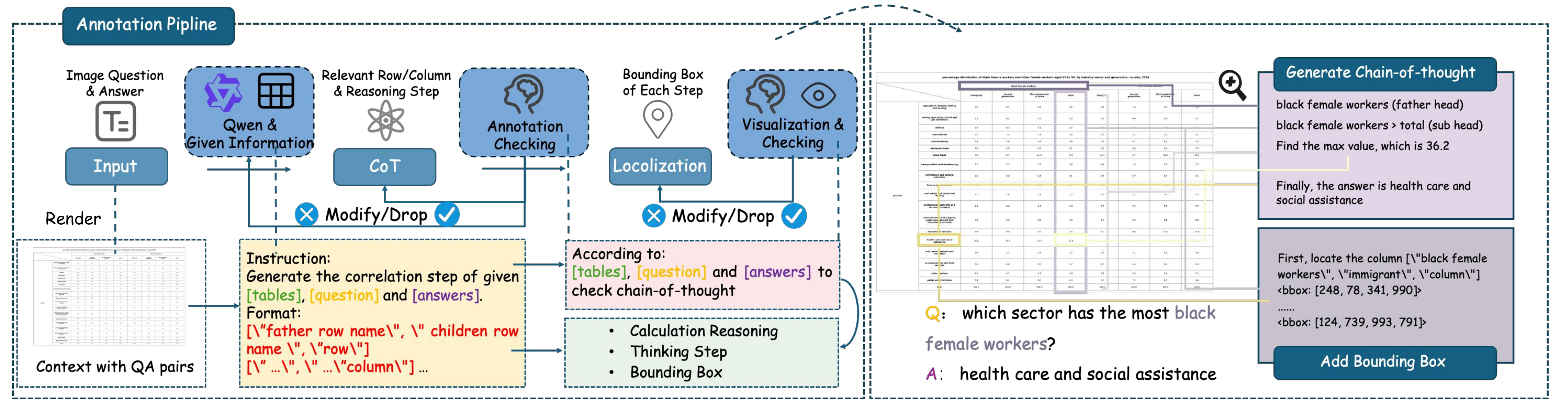}
    \caption{\textbf{The TableVision Annotation Pipeline.} The framework integrates deterministic rendering for coordinate mapping, LLM-based rationale generation (CoT), and a human-in-the-loop verification loop (Modify/Drop) to ensure high-fidelity spatial-logical alignment.}
    \label{fig:pipeline}
\end{figure}

To construct large-scale datasets with precise Bbox and CoT annotations while avoiding coordinate hallucinations, we developed a \textbf{Rendering-Based Deterministic Pipeline}(illustrated in Fig.~\ref{fig:pipeline}).

\paragraph{Step 1: Image Rendering and Coordinate Mapping.}
Starting from the original structured tabular data in JSON format, we programmatically render them into high-resolution images. During this process, the exact pixel coordinates $(x_1, y_1, x_2, y_2)$ of every cell, row header, and column header are automatically logged into a spatial metadata dictionary, ensuring absolute spatial accuracy.

\paragraph{Step 2: Stratified Annotation Generation.}
We apply distinct strategies based on the source of the queries:

\noindent\textbf{Legacy Query Refinement:} For original queries from HiTab, we first perform a rigorous filtering process. Based on the pre-existing answer cell indices, we re-extract the precise Bboxes via the spatial metadata dictionary and employ an LLM to regenerate standardized CoT rationales, ensuring perfect alignment between legacy answers and new visual evidence.

\noindent\textbf{Synthetic Query Generation:} For newly synthesized queries, we utilize an LLM as a logical router to decompose complex queries into \textit{Semantic Tags}. These tags cover 1D array extraction (rows/columns), 2D cell intersection (handling nested headers via the "$>$" operator), and header metadata references.

\paragraph{Step 3: Deterministic Alignment and Decoupling.}
In the final stage, the pipeline executes a deterministic matching algorithm to transform semantic tags into exact Bboxes. Crucially, the CoT rationales and Bbox sets are physically decoupled to prepare the data for training our "Locate-then-Reason" framework. This process transforms implicit reasoning into explicit pairs of \textbf{Spatial Evidence Sets} and \textbf{Logical Chains}, rectifying potential logical drifts in the original data.

\subsection{Quality Assurance and Verification}
\label{subsec:quality_control}

Since any misalignment between a spatial anchor (Bbox) and its corresponding logical step would compromise the benchmark's reliability, we implemented a strict hybrid verification mechanism. First, an automated evaluator parsed the generated reasoning steps, screening for logical hallucinations and spatial boundary violations. Second, any flagged instances, alongside a random 5\% subset of the entire corpus, underwent rigorous human-in-the-loop remediation. 

Crucially, during the validation of the newly synthesized queries, our pipeline identified \textbf{228 instances} exhibiting spatial-logical inconsistencies. For example, a rationale might reference a correct numerical value but remain anchored to an incorrect cell node due to the visual interference of merged headers. These critical topological errors were manually rectified by expert annotators or completely discarded, ensuring the absolute fidelity of the \textit{TableVision} benchmark.

\subsection{Dataset Characteristics and Complexity Analysis}
\label{subsec:characteristics}
To quantify the structural and cognitive density of \textit{TableVision}, we analyze the statistical distributions of the 6,799 instances, as visualized in Fig.~\ref{fig:teaser}(II).

\paragraph{Characterizing the Perceptual Grounding Bottleneck.}
As shown in our Bottleneck Insight (Fig.~\ref{fig:teaser} II-a), an analysis of the dataset metadata reveals a clear disparity between reasoning depth and perceptual demand. Across all 13 categories, the average reasoning length remains relatively stable, ranging from \textbf{4.3 to 4.8 steps}. In contrast, the visual perception burden scales significantly with task complexity: the average number of required spatial anchors increases from \textbf{3.10 Bboxes} in L1 Retrieval to a weighted average of \textbf{10.87 Bboxes} in L3 Analysis, with certain sub-tasks such as Cross-hierarchy Aggregation requiring up to \textbf{17.11 Bboxes}.

\paragraph{Structural Complexity and Linguistic Density.} 
The complexity of \textit{TableVision} is further reflected in its 2D layouts and textual conditions. As shown in (Fig.~\ref{fig:teaser} II-b), the tables exhibit an average dimension of \textbf{14.5 rows $\times$ 9.2 columns}, with an average maximum header depth of \textbf{2.20}. This hierarchical topology poses a challenge to standard localized scanning. Furthermore, each query requires processing an average question length of \textbf{16.9 words} and generating reasoning rationales averaging \textbf{53.8 words}, demanding both robust linguistic processing and spatial alignment.

\paragraph{Evaluation Split and Generalization.} 
To evaluate spatial generalization, we curated the dataset split such that the test set (1,301 queries) exhibits higher multi-modal complexity than the training set (5,498 queries). The average number of spatial anchors increases from \textbf{5.68 Bboxes} in the training set to \textbf{6.24 Bboxes} in the test set, with slightly longer question lengths (\textbf{17.6 vs. 16.7 words}).

\section{Grounding-Enhanced Reasoning Framework}
\label{sec:method}

\subsection{Overview and Problem Formulation}
Given a table image $I$ and a natural language question $Q$, the goal of table reasoning is to generate a correct answer $A$. Standard Multimodal Large Language Models (MLLMs) formulate this directly as maximizing the probability $P(A | I, Q)$. To mitigate the structural entanglement discussed in Section 1, we propose a decoupled framework that explicitly separates spatial perception from logical computation. We reformulate the reasoning process into two sequential stages utilizing intermediate structural bounding boxes $B$:
\begin{equation}
    P(A | I, Q) = \underbrace{P(B | I, Q)}_{\text{Stage 1: Localization}} \times \underbrace{P(A | I, Q, B)}_{\text{Stage 2: Reasoning}}
\end{equation}

Our framework is built upon the Qwen3-VL-8B-Instruct backbone. To achieve parameter-efficient adaptation, we apply Low-Rank Adaptation (LoRA) to the language model while keeping the vision tower and vision-language projector frozen. The training is implemented using the LLaMA-Factory framework.

\subsection{Stage 1: Explanatory Structural Grounding}
The objective of the first stage is to identify the relevant row headers, column headers, and target cells necessary for answering the query.

\noindent\textbf{Input Formulation and Tokenization.} The model acts as a structural parser guided by a system prompt. The prompt defines five semantic label types (column, row, cell, colhead, and rowhead) to navigate the 2D topology. To map continuous visual space to text, we employ a 1000-level normalization scheme, representing coordinates as discrete integers in the range $[0, 999]$.

\noindent\textbf{Autoregressive Generation.} For a given query, the model first generates a brief textual explanation (Reason) justifying its focus. Following this rationale $R$, it autoregressively outputs the localized bounding boxes in the format \texttt{[label] (x1,y1)(x2,y2)}. The sequence termination is naturally handled by the end-of-sentence token. We optimize this stage using the standard next-token prediction loss over the rationale and bounding box tokens $B$:
\begin{equation}
    \mathcal{L}_{loc} = - \sum_{t=1}^{|B|} \log P_{\theta}(b_t | I, Q, R, b_{<t})
\end{equation}
\begin{figure}[t]
    \centering
    \includegraphics[width=1\linewidth]{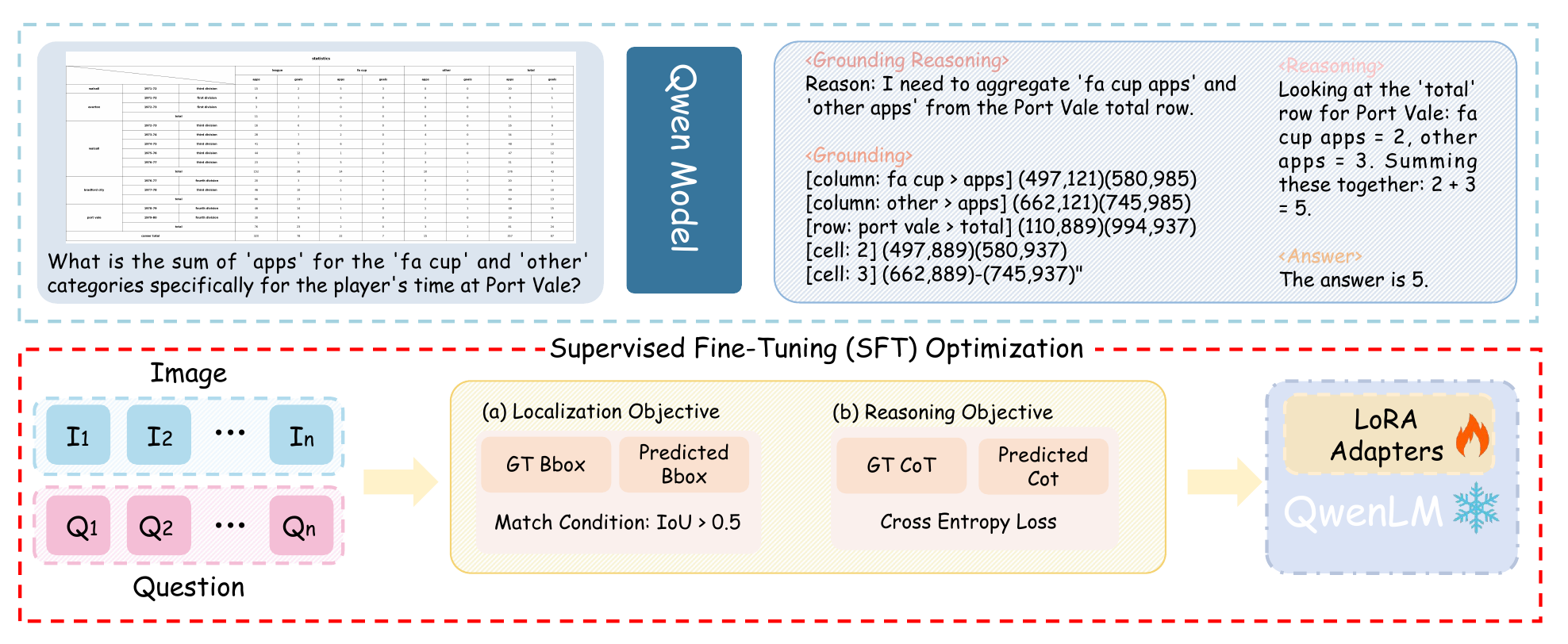} 
    \caption{The SFT training pipeline of our framework. We fine-tune LoRA adapters on a frozen Qwen3-VL-8B-Instruct backbone. The process is decoupled into Explanatory Structural Localization and Grounding-Conditioned Reasoning.}
    \label{fig:training}
\end{figure}
\subsection{Stage 2: Grounding-Conditioned Reasoning}
\label{subsec:stage2}

In the second stage, the model performs logical synthesis using the localized regions as explicit structural evidence. 

\noindent\textbf{Evidence-Reasoning Integration.} The coordinates and semantic labels generated in Stage 1 are integrated as text tokens into the prompt for Stage 2. This explicit anchoring forces the self-attention mechanism to prioritize specific visual patches during the reasoning process. By providing these spatial anchors, we decouple the cognitive burden: the model's parameter capacity is dedicated to numerical computation and cross-hierarchical logic rather than visual spatial searching.

\noindent\textbf{Reasoning Objective.} The model is trained to generate a Chain-of-Thought (CoT) $C$ that cites the grounded coordinates before producing the final answer $A$. Given the input image $I$, query $Q$, and the predicted spatial anchors $\hat{B}$, the training objective for this stage is formulated as:
\begin{equation}
    \mathcal{L}_{reason} = - \sum_{t=1}^{|C, A|} \log P_{\phi}(c_t | I, Q, \hat{B}, c_{<t})
\end{equation}
where $\phi$ denotes the trainable parameters of the LoRA adapters applied to the language model. This conditional generation ensures that the resulting reasoning trajectory is physically anchored to the structural evidence identified in the first stage.

\section{Experiments}
\label{sec:experiments}

In this section, we conduct extensive experiments to: (1) Benchmark the zero-shot performance of 10 Multimodal Large Language Models (MLLMs) on the \textit{TableVision} dataset; (2) Utilize an Oracle probing strategy to isolate and diagnose the perception bottleneck; and (3) Evaluate the effectiveness of our proposed two-stage "Locate-then-Reason" framework.

\subsection{Experimental Setup}
\label{subsec:setup}

\textbf{Baselines for Benchmarking.} To assess current capabilities in hierarchical table reasoning, we evaluate 10 representative MLLMs with active parameter scales ranging from 4B to 9B. The selection includes general-purpose models (e.g., MiniCPM-V-2.6\cite{yao2024minicpm}, InternVL3-8B\cite{zhu2025internvl3}, Qwen3-VL-8B-Instruct\cite{bai2025qwen3vltechnicalreport}) and table-specialized models (e.g., Glyph\cite{cheng2025glyphscalingcontextwindows}). All models are evaluated in a zero-shot setting to measure their inherent structural perception. These standard VLMs perform inference by directly generating the final answer without extracting bounding boxes.

\textbf{Implementation Details for Our Framework.} For our two-stage framework, we utilize \textbf{Qwen3-VL-8B-Instruct} as the base model. The Supervised Fine-Tuning (SFT) is conducted using the LLaMA-Factory framework. We employ LoRA on all linear layers of the language model with a rank of $r=32$ and an alpha of $\alpha=64$, while the visual encoder and projector remain frozen. To ensure reproducibility, both Stage 1 (Localization) and Stage 2 (Reasoning) are trained using the identical hyperparameter configuration. Specifically, the model is trained for 2 epochs using the AdamW optimizer with a learning rate of $1 \times 10^{-5}$ and a cosine decay schedule. The total batch size is set to 32 across 4 NVIDIA A100 (40GB) GPUs.

\textbf{Evaluation Metrics.} We report the exact-match \textbf{Accuracy (\%)} across the 13 fine-grained task categories spanning cognitive levels L1 to L3.

\subsection{Benchmarking Models}
\label{subsec:sota_results}

Table~\ref{tab:benchmark_results} presents the comprehensive zero-shot evaluation of the 10 MLLMs on the \textit{TableVision} test set. The results reveal several critical insights into the current limitations of multimodal architectures.

\textbf{Capability Degradation Across Levels.} While most models perform adequately on L1 Perception tasks (e.g., Qwen3-VL achieves 78.8\% on Retrieval), their performance degrades significantly as cognitive complexity increases. On L3 Analysis tasks, nearly all general-purpose models suffer severe performance drops, with many falling below 30\% accuracy (e.g., Qwen3-VL drops to 24.3\% on Compositional Arithmetic).

\textbf{Challenge of Hierarchical Topologies.} Even table-specialized models struggle to maintain high accuracy across all tasks. Glyph achieves the highest overall accuracy (71.6\%), but still exhibits weaknesses in complex multi-hop scenarios. This confirms that \textit{TableVision} poses a significant challenge as a benchmark for multimodal architectures.

\begin{table*}[t]
\centering
\caption{\textbf{Comprehensive Zero-Shot Evaluation on the TableVision Test Set.} Accuracy is reported in percentage (\%). We evaluate 10 VLMs with comparable active parameter scales (4B to 9B). Evaluated variants include: MiniCPM-V-2.6, HiPPO, InternVL3-8B, Phi-3.5-Vision-Instruct, LLaVA-OneVision-7B, DeepSeek-VL2 (27B MoE, 4.5B active), GLM-4.1V-9B, Ovis2.5-9B, Glyph(GLM-4.1V-9B-Base), and Qwen3-VL-8B-Instruct.}
\label{tab:benchmark_results}
\resizebox{\textwidth}{!}{%
\begin{tabular}{@{} l c | cccccccccc @{}}
\toprule
\textbf{Category} & \textbf{\textit{N}} & \textbf{MiniCPM-V} & \textbf{HiPPO} & \textbf{InternVL3} & \textbf{Phi-3.5-V} & \textbf{LLaVA-OV} & \textbf{DS-VL2} & \textbf{GLM-4.1V} & \textbf{Ovis2.5} & \textbf{Glyph} & \textbf{Qwen3-VL} \\
\textit{Params} & - & \textit{8B} & \textit{8B} & \textit{8B} & \textit{4.2B} & \textit{7B} & \textit{4.5B*} & \textit{9B} & \textit{9B} & \textit{9B} & \textit{8B} \\
\midrule
\multicolumn{12}{@{}l}{\textbf{L1: Perception (N=513)}} \\
Retrieval & 371 & 73.0 & 73.6 & 36.4 & 24.8 & 57.7 & 61.2 & 80.6 & 76.5 & 78.4 & 78.8 \\
Listing & 28 & 57.1 & 39.3 & 25.0 & 21.4 & 21.4 & 17.9 & 42.9 & 75.0 & 82.1 & 59.3 \\
Structure & 114 & 66.7 & 39.5 & 48.2 & 20.2 & 50.0 & 61.4 & 41.2 & 20.2 & 78.1 & 90.4 \\
\midrule
\multicolumn{12}{@{}l}{\textbf{L2: Reasoning (N=425)}} \\
Comparison & 113 & 58.4 & 65.5 & 27.4 & 10.6 & 21.2 & 36.3 & 53.1 & 61.9 & 58.4 & 62.3 \\
Arithmetic & 48 & 54.2 & 60.4 & 20.8 & 4.2 & 29.2 & 50.0 & 75.0 & 68.8 & 60.4 & 77.6 \\
Ranking & 47 & 59.6 & 53.2 & 31.9 & 19.1 & 36.2 & 46.8 & 70.2 & 70.2 & 72.3 & 70.5 \\
Cond. Filtering & 13 & 30.8 & 23.1 & 30.8 & 23.1 & 38.5 & 23.1 & 61.5 & 61.5 & 92.3 & 61.5 \\
Counting & 92 & 37.0 & 39.1 & 26.1 & 31.5 & 25.0 & 34.8 & 40.2 & 58.7 & 67.4 & 41.3 \\
Verification & 112 & 59.8 & 58.9 & 60.7 & 50.9 & 53.6 & 48.2 & 66.1 & 82.1 & 92.0 & 65.2 \\
\midrule
\multicolumn{12}{@{}l}{\textbf{L3: Analysis (N=363)}} \\
Comp. Arith. & 107 & 10.3 & 12.1 & 5.6 & 6.5 & 3.7 & 5.6 & 13.1 & 66.4 & 78.5 & 24.3 \\
Cross-hier. Agg. & 72 & 13.9 & 13.9 & 5.6 & 5.6 & 8.3 & 1.4 & 16.7 & 52.8 & 66.7 & 27.8 \\
Multi-hop & 104 & 6.7 & 4.8 & 1.9 & 4.8 & 8.7 & 10.6 & 13.5 & 25.0 & 58.7 & 26.0 \\
Temporal & 80 & 12.5 & 23.8 & 17.5 & 12.5 & 13.8 & 13.8 & 21.2 & 16.2 & 37.5 & 32.5 \\
\midrule
\textbf{Overall} & \textbf{1301} & 48.1 & 46.8 & 28.8 & 19.9 & 34.6 & 39.0 & 51.0 & 58.9 & 71.6 & 59.3 \\
\bottomrule
\end{tabular}%
}
\end{table*}

\subsection{Diagnosing the Perception Bottleneck}
\label{subsec:oracle_probing}
To empirically validate the "Perception Bottleneck" hypothesis, we first evaluate the standalone performance of our Stage-1 localization module and its direct impact on downstream reasoning.

\begin{figure}[t]
\centering
\small
\begin{tabular}{cc}
\begin{tabular}[b]{c}
    \begin{tabular}{c|c|cc|ccc}
        \toprule
        \textbf{Ckpt} & \textbf{Pairs} & \textbf{Mean} & \textbf{Median} & \textbf{$\ge$0.5} & \textbf{$\ge$0.75} & \textbf{$\ge$0.9} \\
        \midrule
        50  & 2202 & 0.287 & 0.198 & 24.9\% & 7.0\%  & 1.4\%  \\
        100 & 2847 & 0.585 & 0.657 & 67.2\% & 36.4\% & 9.3\%  \\
        150 & 2453 & 0.527 & 0.648 & 60.3\% & 36.4\% & 11.3\% \\
        200 & 2458 & 0.542 & 0.672 & 61.8\% & 39.4\% & 12.2\% \\
        \bottomrule
    \end{tabular} \\
    \footnotesize (a) Localization Metrics
\end{tabular} & 

\begin{tabular}[b]{c}
    \raisebox{-20pt}{\includegraphics[height=2.4cm]{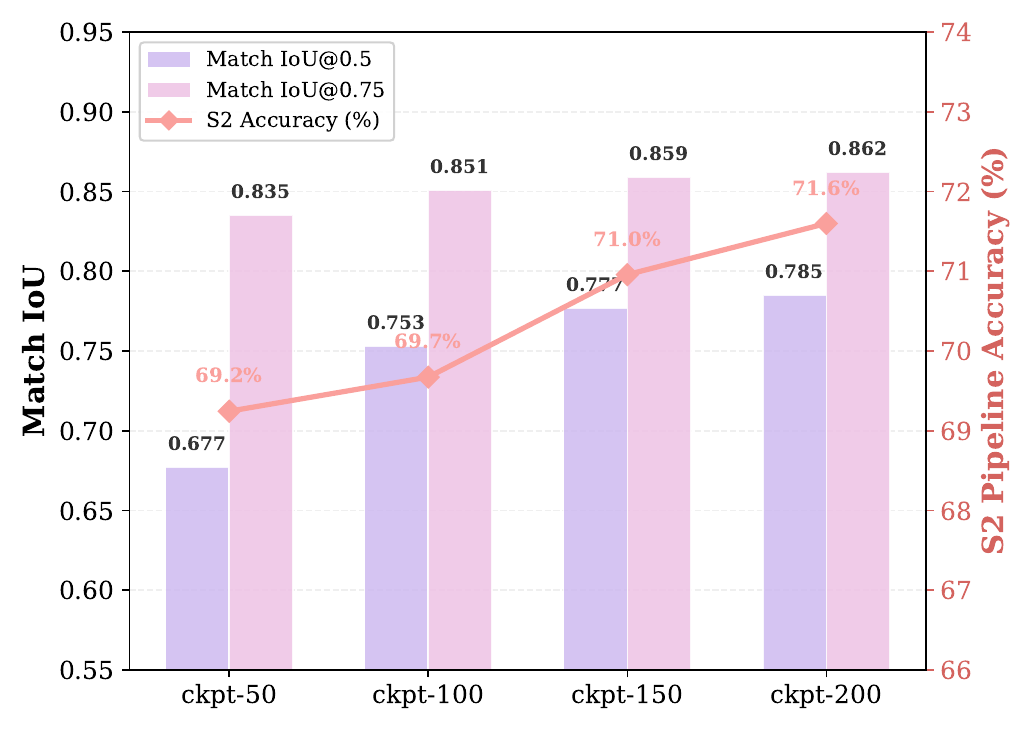}} \\
    \footnotesize (b) Grounding vs. Accuracy
\end{tabular}
\end{tabular}

\caption{Quantitative analysis of Stage-1 spatial grounding and its correlation with downstream S2 Pipeline accuracy.}
\label{fig:combined_analysis}
\end{figure}

\paragraph{Stage-1 Localization Performance.}
As shown in Fig.~\ref{fig:combined_analysis}a), we report the localization metrics of Stage-1 across training checkpoints. At Ckpt 200, the model achieves a median IoU of \textbf{0.672}, with 61.8\% of predicted boxes exceeding the 0.5 IoU threshold. However, the performance drops significantly as the precision requirement increases, with only \textbf{12.2\%} of boxes achieving \textbf{IoU $\ge$ 0.9}. This confirms that high-precision localization in dense tables remains the primary challenge.

\paragraph{Localization Quality vs. Downstream Accuracy.}
We further analyze the correlation between grounding precision and final task accuracy. As illustrated in Fig.~\ref{fig:combined_analysis}(b), the downstream reasoning accuracy is highly sensitive to Stage-1 quality. When the localization median IoU improves from 0.198 to 0.672, the overall S2 Pipeline accuracy increases accordingly from \textbf{69.2\% to 71.6\%}. This empirical correlation validates the "Locate-then-Reason" paradigm: stronger spatial anchoring directly translates to more robust logical synthesis.

\paragraph{Oracle Probing.} 
To isolate the reasoning bottleneck, we provide the baseline with ground-truth (GT) bounding boxes. Removing the spatial search burden surges overall accuracy by \textbf{20.7\%} (59.3\% $\to$ 80.0\%). This leap is most evident in L3 tasks; e.g., Compositional Arithmetic improves by \textbf{65.4\%} (24.3\% $\to$ 89.7\%). These results prove the model possesses the requisite logic but fails in zero-shot settings due to spatial disorientation—operating on incorrect semantic content.

\paragraph{Error Propagation Analysis.} 
While our Decoupled SFT yields a 12.3\% overall gain, it marginally underperforms the End-to-End baseline on L3 tasks (59.2\% vs. 60.6\%). This exposes the vulnerability of Stage-1: L3 queries require dense spatial anchors (avg. \textbf{10.87 Bboxes}), leading to error accumulation during auto-regressive localization that misguides Stage-2. Nonetheless, the Oracle L3 accuracy of \textbf{67.5\%} confirms that precise grounding remains the ultimate bottleneck.

\paragraph{Performance Regression in L1.}
We observe an accuracy drop in L1 tasks following SFT (Table~\ref{tab:ablation_results}). This regression likely stems from: 
(1) \textbf{Reasoning Overhead}: While Zero-shot models leverage direct visual attention for simple retrieval, our SFT models are mandated to follow a structured CoT. This added complexity can introduce stochastic errors in low-level tasks compared to direct extraction. 
(2) \textbf{Task Interference}: The optimization for complex logic in L2 and L3 may overshadow simple perceptual shortcuts, a common trade-off in multi-task fine-tuning.

\begin{table}[t]
\centering
\caption{Ablation Study on the TableVision Test Set. \textbf{Zero-shot}: Qwen3-VL-8B baseline. \textbf{End-to-End SFT}: Single-stage supervised model predicting CoT and answer directly. \textbf{Decoupled SFT (Ours)}: Our pipeline predicting bounding boxes before CoT. \textbf{Oracle}: Setting using ground-truth boxes.}
\label{tab:ablation_results}
\begin{tabular}{l|ccc|c}
\toprule
\textbf{Setting} & \textbf{Perception} & \textbf{Reasoning} & \textbf{Analysis} & \textbf{Overall} \\
\midrule
\# Test Samples & 513 & 425 & 363 & 1301 \\
\midrule
Zero-shot & 80.3 & 61.1 & 27.3 & 59.3 \\
End-to-End SFT & 70.8 {\scriptsize (\textcolor{red}{-9.5})} & 61.6 {\scriptsize (\textcolor{teal}{+0.5})} & 60.6 {\scriptsize (\textcolor{teal}{+33.3})} & 65.0 {\scriptsize (\textcolor{teal}{+5.7})} \\
Decoupled SFT (Ours) & 79.1 {\scriptsize (\textcolor{red}{-1.2})} & 73.1 {\scriptsize (\textcolor{teal}{+12.0})} & 59.2 {\scriptsize (\textcolor{teal}{+31.9})} & 71.6 {\scriptsize (\textbf{\textcolor{teal}{+12.3}})} \\
Oracle (GT Bbox) & 85.8 {\scriptsize (\textcolor{teal}{+5.5})} & 83.9 {\scriptsize (\textcolor{teal}{+22.8})} & 67.5 {\scriptsize (\textcolor{teal}{+40.2})} & 80.0 {\scriptsize (\textbf{\textcolor{teal}{+20.7}})} \\
\bottomrule
\end{tabular}
\end{table}

\subsection{Sensitivity and Evidence Utilization Analysis}
\label{subsec:sensitivity}

\paragraph{Accuracy vs. Number of Spatial Anchors.}
We evaluate how reasoning accuracy fluctuates as the density of spatial anchors increases. As the required boxes grow from sparse (1 to 5) to dense (greater than 10), the zero-shot baseline experiences a sharp performance drop. In contrast, our decoupled framework maintains a significantly higher accuracy across the same density intervals, indicating that explicit spatial grounding buffers the model against perceptual overload.

\paragraph{Validation of Evidence Utilization via Oracle Performance.}
To address concerns regarding whether the model genuinely relies on provided spatial evidence, we identify the \textbf{Oracle performance leap} as the most direct metric of evidence utilization. As demonstrated in Table~\ref{tab:ablation_results}, providing precise spatial anchors (Oracle) enables the reasoning module to achieve a massive +40.2\% accuracy gain on L3 tasks compared to the zero-shot baseline. This substantial improvement quantitatively proves that the Stage-2 module possesses a high degree of \textbf{evidence faithfulness}: once the perception layer (Stage-1) provides accurate spatial coordinates, the reasoning module can faithfully synthesize logical answers based on the anchored visual content rather than falling back on hallucinations.

\section{Conclusion}
\label{sec:conclusion}

In this paper, we investigate the structural entanglement problem faced by Multimodal Large Language Models (MLLMs) when processing complex hierarchical tables. We identify a critical Perception Bottleneck through quantitative analysis, revealing that as task complexity scales, MLLMs struggle to maintain accurate spatial attention amidst increasing visual region density. To mitigate this, we introduce TableVision, a benchmark dataset comprising 6,799 high-fidelity instances. By providing explicit bounding box trajectories and Chain-of-Thought annotations, our dataset supports the decoupling of dense tabular reasoning into a "Locate then Reason" paradigm. Despite these advancements, this study has certain limitations. The final reasoning performance of our two-stage framework remains heavily constrained by the localization accuracy of the first stage; specifically, L3 Analysis tasks, which require an average of 10.87 spatial anchors, are particularly vulnerable to error propagation during the auto-regressive grounding phase. Additionally, because existing tabular datasets typically lack fine-grained spatial coordinate ground truths, direct cross-dataset generalization evaluation is currently restricted. Future work will focus on improving the structural parsing robustness of models on extremely dense tables and extending this decoupled paradigm to broader Visually-Rich Document Understanding (VrDU) tasks to bridge the gap between perceptual grounding and high-level logical synthesis.

\bibliographystyle{splncs04}
\bibliography{main}
\end{document}